\documentclass{article}

\usepackage{microtype}
\usepackage{graphicx}
\usepackage{subcaption}
\usepackage{booktabs}
\usepackage{multirow}
\usepackage{hyperref}
\usepackage{float} 
\usepackage{stfloats}

\usepackage[preprint]{icml2026}

\hypersetup{pdfsubject={}}

\usepackage{amsmath, amssymb, amsfonts}
\usepackage{bm}
\usepackage{amsthm}
\usepackage{xspace}
\usepackage{enumitem}

\newtheorem{assumption}{Assumption}

\newcommand{\eg}{e.g.,\xspace}

\DeclareMathOperator*{\argmax}{arg\,max}

\icmltitlerunning{ScaleToT: Generalizing Structured LLM Reasoning for Billion-Scale Low-Activity User Modeling}

\begin{document}
\raggedbottom

\twocolumn[
\icmltitle{ScaleToT: Generalizing Structured LLM Reasoning for\\ Billion-Scale Low-Activity User Modeling}

\icmlsetsymbol{equal}{*}

\begin{icmlauthorlist}
\icmlauthor{Tianbao Ma}{equal,ks}
\icmlauthor{Chang Xi}{equal,ks}
\icmlauthor{Yichuan Zou}{ks}
\icmlauthor{Chengen Li}{ks}
\icmlauthor{Linxun Chen}{ks}
\icmlauthor{Zilong Lu}{ks}
\icmlauthor{Yanan Niu}{ks}
\icmlauthor{Zhaojie Liu}{ks}
\icmlauthor{Han Li}{ks}
\icmlauthor{Kun Gai}{ks}
\end{icmlauthorlist}

\icmlaffiliation{ks}{Kuaishou Technology, Beijing, China}

\icmlcorrespondingauthor{Linxun Chen}{chenxi36@kuaishou.com}

\icmlkeywords{User Modeling, Large Language Models, Structured Reasoning, Tree-of-Thought, Lifetime Value Prediction}

\vskip 0.3in
]

\printAffiliationsAndNotice{\icmlEqualContribution}

\begin{abstract}
Accurate user modeling often depends on rich interaction histories, which are unavailable for billions of low-activity users. Large Language Models (LLMs) can infer latent user states from static profiles, but this reasoning becomes unreliable when profiles are sparse, and applying an LLM to billions of users is prohibitively expensive. We present ScaleToT, which learns structured reasoning from a small LLM-processed subset and extends it to the broader low-activity user population. To improve reasoning reliability, ScaleToT constructs typed user-state chains with a bounded entropy-guided Tree-of-Thought (ToT) refinement procedure. To make this structured reasoning usable from sparse profiles, the teacher-curated chains are used to train a student model on static profiles through supervised fine-tuning (SFT) and Outcome-Driven Segment-Aware Implicit Reward Policy Optimization (OSIPO). ScaleToT then transfers the student's reasoning representations to a lightweight profile encoder, providing shared reasoning signals for the remaining users without LLM inference. We evaluate ScaleToT on lifetime value (LTV) prediction in a billion-scale advertising deployment. A randomized online A/B test increased LT30 by 6.738\%, while offline reasoning covered only 7.32\% of the potential population, greatly reducing compute cost compared with full-population reasoning.
\end{abstract}

\section{Introduction}\label{sec:intro}

User understanding is a shared foundation for recommendation, personalization, and user-growth platforms at billion scale~\citep{he2023survey}, aiming to infer the latent states and intentions that drive behavior rather than merely fitting surface interaction patterns~\citep{chang2023latent}. The dominant paradigm casts this as behavior-sequence modeling over interaction streams such as clicks, purchases, and browsing events~\citep{zhang2025delrec,xia2022multi}, and excels precisely when each user supplies a sufficient, recent, and information-rich history~\citep{zhai2024actions}. A large and commercially critical population violates this assumption at its core: low-activity users, and in particular dormant users targeted by advertising customer acquisition campaigns, leave histories too sparse or too stale to serve as reliable behavioral evidence~\citep{monteil2024marec,li2022transform,zhang2022diverse,li2022billion}. For these users the industrial objective also shifts, from predicting the next item to estimating lifetime value (LTV), retention potential, and advertising yield upon re-engagement~\citep{su2023cross,wang2024adsnet}, conditioned almost entirely on a sparse static profile~\citep{yang2023feature}. Existing remedies for sparsity, including auxiliary-information fusion and cross-domain transfer, still consume structured behavioral signals as their primary input~\citep{li2022transform,xiao2025mars}, which is exactly what these users do not provide.

Our starting point is a reframing of what this regime actually demands. Behavioral sparsity does not imply that a user cannot be understood~\citep{li2023stan}; even a dormant user carries latent preferences and re-engagement potential weakly reflected in static profile cues~\citep{liu2024modeling}. The difficulty is that these states must be inferred from sparse evidence rather than observed behavior. The task therefore shifts from extrapolating the next interaction to constructing a reliable user representation from static profiles. This is where Large Language Models (LLMs) are useful: not as larger sequence encoders, but as reasoners that apply world knowledge about user motivations and behavioral drivers when behavioral traces are absent~\citep{xi2024towards,wang2024llmrg,wang2025lettingo,zhai2025choirrec}.

\begin{figure}[t]
  \centering
  \begin{subfigure}{0.49\linewidth}
    \centering
    \includegraphics[width=\linewidth]{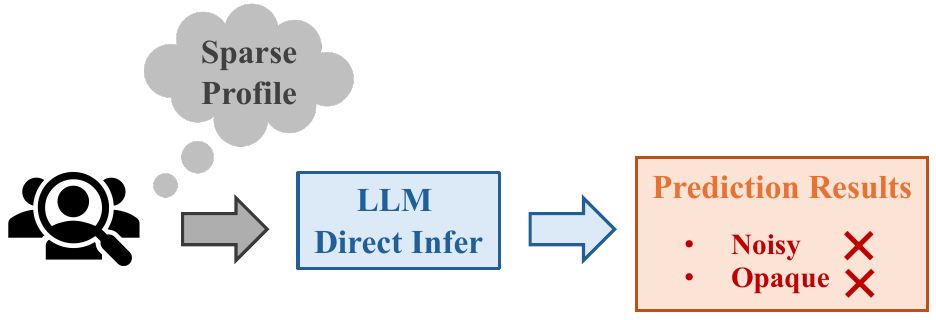}
    \caption{}
    \label{fig:motiv-sparse}
  \end{subfigure}
  \hfill
  \begin{subfigure}{0.49\linewidth}
    \centering
    \includegraphics[width=\linewidth]{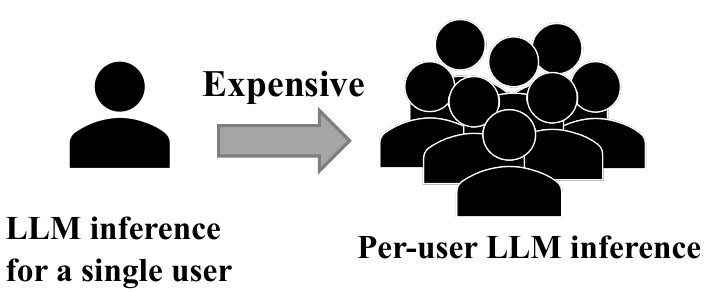}
    \caption{}
    \label{fig:motiv-cost}
  \end{subfigure}
  \caption{Two obstacles to applying LLM reasoning to low-activity users. (a) On a sparse profile, a direct LLM query gives a noisy and opaque prediction with no inspectable intermediate state. (b) Running one LLM inference per user makes cost grow with the user base. Together they make LLM reasoning over a billion-scale population both unreliable and unaffordable.}
  \label{fig:intro}
\end{figure}

Direct LLM reasoning in this setting faces two challenges. First, a single prediction from a sparse profile collapses several latent dimensions into an opaque judgment, leaving no intermediate state to inspect, constrain, or refine. Second, per-user reasoning makes production cost grow linearly with the target population. Thus, useful LLM reasoning must be made both reliable under sparse evidence and reusable beyond the users on which it is generated.

ScaleToT learns structured reasoning from a small offline subset and applies it to population-scale prediction from sparse profiles. Training examples may include privileged information that is unavailable at inference time. Because sparse profiles do not directly reveal task-relevant latent states, a teacher uses this information within a bounded Tree-of-Thought (ToT) procedure to construct outcome-consistent reasoning chains. The procedure maintains parallel interpretations of typed user states, evaluates their uncertainty, and selectively revises uncertain fields rather than regenerating an entire chain. The resulting chains serve as training data for a student model that reasons from static profiles through supervised fine-tuning (SFT) and our Outcome-Driven Segment-Aware Implicit Reward Policy Optimization (OSIPO). ScaleToT then aligns a lightweight profile encoder with student reasoning and uses vector-quantized (VQ) codebooks to retrieve shared reasoning patterns, enabling population-scale prediction without LLM inference. We formulate ScaleToT generally but evaluate it only on low-activity user LTV prediction, where five typed fields capture lifecycle dimensions.

The main contributions of this work are summarized as follows:
\begin{itemize}
  \item We make LLM-based user-state reasoning reliable under sparse profiles by representing latent user states as typed chains and refining them with a bounded entropy-guided ToT refinement procedure. The procedure preserves plausible alternatives and locally revises uncertain states, while OSIPO provides segment-aware, outcome-driven supervision without requiring state-level annotations.

  \item We make structured reasoning reusable at population scale by aligning a lightweight profile encoder with student-generated chains and quantizing recurring reasoning patterns into VQ codebooks. At inference time, users retrieve profile-conditioned reasoning prototypes without LLM generation or chain embedding.

  \item We validate ScaleToT in a billion-scale advertising deployment for LTV prediction. Structured reasoning is generated for only 7.32\% of the potential population, while a randomized online A/B test improves LT30 by 6.738\%.
\end{itemize}

\section{Related Work}\label{sec:related}
\paragraph{Low-Activity User Modeling.}
Existing methods compensate for sparse interactions with metadata~\citep{gantner2010learning,monteil2024marec}, meta-learning~\citep{vartak2017meta,moghaddam2026user}, bandit exploration~\citep{nguyen2014cold}, or multimodal evidence~\citep{pan2022multimodal}. Others transfer information across users, activity regimes, or domains~\citep{zhu2021cross}: Cold-Transformer~\citep{li2022transform} adapts representations across activity levels, MARS~\citep{xiao2025mars} retrieves high-activity users to enrich sparse histories, and cross-domain LTV methods learn invariant representations~\citep{su2023cross} or augment samples with external data~\citep{wang2024adsnet}. These methods remain anchored in observed behaviors or behavior-linked signals, making them less effective when histories are nearly absent or stale.

\paragraph{LLMs for User Modeling and Recommendation.}
LLMs can provide external knowledge or explicit reasoning for recommendation. KAR~\citep{xi2024towards} elicits preference and factual knowledge to enrich conventional representations, while LLMRG~\citep{wang2024llmrg} constructs and verifies personalized reasoning graphs from profiles and behavioral sequences. More generally, chain-of-thought (CoT) prompting~\citep{wei2022chain}, Tree-of-Thought (ToT) search~\citep{yao2023tree}, process reward models (PRMs)~\citep{lightman2024let}, and reinforcement learning (RL) with verifiable rewards~\citep{guo2025deepseek} expose or supervise intermediate reasoning. Recent methods such as LettinGo~\citep{wang2025lettingo}, EXP3RT~\citep{kim2025driven}, ELEC~\citep{dong2025elec}, and ChoirRec~\citep{zhai2025choirrec} apply these capabilities to user representation, but do not jointly address unreliable reasoning from near-empty profiles and population-scale inference cost.

Efficiency-oriented systems pre-compute interest-cluster transitions~\citep{wang2024llms} or separate LLM processing from inference~\citep{xi2025efficient}, yet still derive reusable structures primarily from behavioral histories. ScaleToT instead instantiates ToT as bounded typed search over a finite set of task-defined state fields rather than open-ended tree expansion, learns from training-only privileged supervision, transfers reasoning at the population level through shared VQ prototypes rather than per-user distillation, and performs inference from profiles alone. Entropy ranks refinement priority rather than estimating calibrated error, and the evaluated sampling ratio is specific to our deployment setting.

\section{Problem Formulation}\label{sec:preliminaries}

\subsection{General Sparse-Input Structured Reasoning Transfer}
We consider sparse-input prediction with training-only privileged context. Let $\mathcal{U}$ be the full population and $\mathcal{U}_{\mathrm{sp}}\subset\mathcal{U}$ the subset whose inference-time inputs are too sparse for sequential modeling. Each instance $u\in\mathcal{U}_{\mathrm{sp}}$ has a sparse input $\bm{x}_u\in\mathbb{R}^{d}$ available by the prediction cutoff, a target $y_u\in\mathcal{Y}$, and optional privileged context $p_u$ available only for historical supervision.

We represent the latent state underlying $y_u$ by $K$ task-defined typed fields:
\begin{equation}\label{eq:chain}
  \begin{aligned}
    \bm{c}_u&=(c_u^{1},\dots,c_u^{K}),\\
    c_u^{k}&\in\mathcal{T}_k=\{t_{k,m}\}_{m=1}^{|\mathcal{T}_k|},\quad k=1,\dots,K,
  \end{aligned}
\end{equation}
where $c_u^{k}$ is the value of field $k$, $\mathcal{T}_k$ its finite candidate set, and $t_{k,m}$ its $m$-th candidate. Fields are generated independently and serialized in a fixed order; this order implies neither temporal nor causal dependence, and $|\mathcal{T}_k|$ may vary across fields.

ScaleToT applies under the following conditions.
\begin{assumption}\label{asm:applic}
\emph{(i)}~latent task-relevant states can be represented by a finite set of typed fields; \emph{(ii)}~training instances admit optional privileged supervision; and \emph{(iii)}~per-instance LLM inference is infeasible or undesirable in the prediction path.
\end{assumption}

\subsection{Prediction-Time Information Boundary}
For each instance, $t_0$ is the prediction cutoff. The input $\bm{x}_u$ contains only information available by $t_0$; privileged context $p_u$ may annotate training examples but is excluded from the student, input encoder, codebook query, validation/test data, and inference request.

\subsection{ScaleToT Objective}
Because per-instance LLM reasoning is infeasible at scale, it is restricted to a historical subset $\mathcal{U}_{\mathrm{LLM}}\subset\mathcal{U}_{\mathrm{sp}}$ with ratio $\rho=|\mathcal{U}_{\mathrm{LLM}}|/|\mathcal{U}_{\mathrm{sp}}|$.
Let $\tilde{\bm{c}}_u$ be the sparse-input student chain, $f_{\mathrm{llm}}$ a fixed function that produces its reasoning embedding, $f_{\mathrm{user}}$ the input encoder, $f_\theta$ the predictor, and $\Psi_\theta$ the VQ retrieval and fusion function. We optimize:
\begin{subequations}\label{eq:problem}
\begin{align}
  \min_{f_\theta,\,f_{\mathrm{user}},\,\Psi_\theta}\quad
  & \sum_{u\in\mathcal{U}_{\mathrm{sp}}}\mathcal{L}_{\mathrm{cal}}\!\bigl(f_\theta(\bm{x}_u,\bm{e}_u^{\mathrm{agg}}),y_u\bigr)\\
  \mathrm{s.t.}\quad
  & \mathrm{C1}:\ c_u^{k}\in\mathcal{T}_k,\quad u\in\mathcal{U}_{\mathrm{LLM}},\ k=1,\ldots,K,\\
  & \mathrm{C2}:\ \bm{z}_u=
    \begin{cases}
      f_{\mathrm{llm}}(\tilde{\bm{c}}_u), & u\in\mathcal{U}_{\mathrm{LLM}},\\
      f_{\mathrm{user}}(\bm{x}_u), & u\notin\mathcal{U}_{\mathrm{LLM}},
    \end{cases}\\
  & \mathrm{C3}:\ \bm{e}_u^{\mathrm{agg}}=\Psi_\theta(\bm{x}_u),\quad u\in\mathcal{U}_{\mathrm{sp}}.
\end{align}
\end{subequations}
Here $\mathcal{L}_{\mathrm{cal}}$ is cross-entropy loss. C1 constrains every field, C2 aligns student-reasoning and sparse-input representations, and C3 makes the deployed representation a function of $\bm{x}_u$ alone. Thus, privileged evidence shapes offline supervision on $\mathcal{U}_{\mathrm{LLM}}$ but is unavailable when the model is applied to $\mathcal{U}_{\mathrm{sp}}$.

\subsection{Advertising LTV Instantiation}
For advertising delivery scenario, $\bm{x}_u$ contains demographics, device attributes, tenure, registration metadata, and pre-cutoff statistics; $y_u$ is a High/Medium/Low LTV activity-intensity label. The $K=5$ fields are user group, churn reason, return motivation, interest preference, and behavior pattern, while $p_u$ contains conversion context and post-return feedback and statistics. Users inactive for at least 30 consecutive days are treated as low-activity users, and their activity during the seven days after return is used to construct the LTV label.

\section{Methods}\label{sec:method}
\begin{figure*}[t!]
  \centering
  \includegraphics[width=\linewidth]{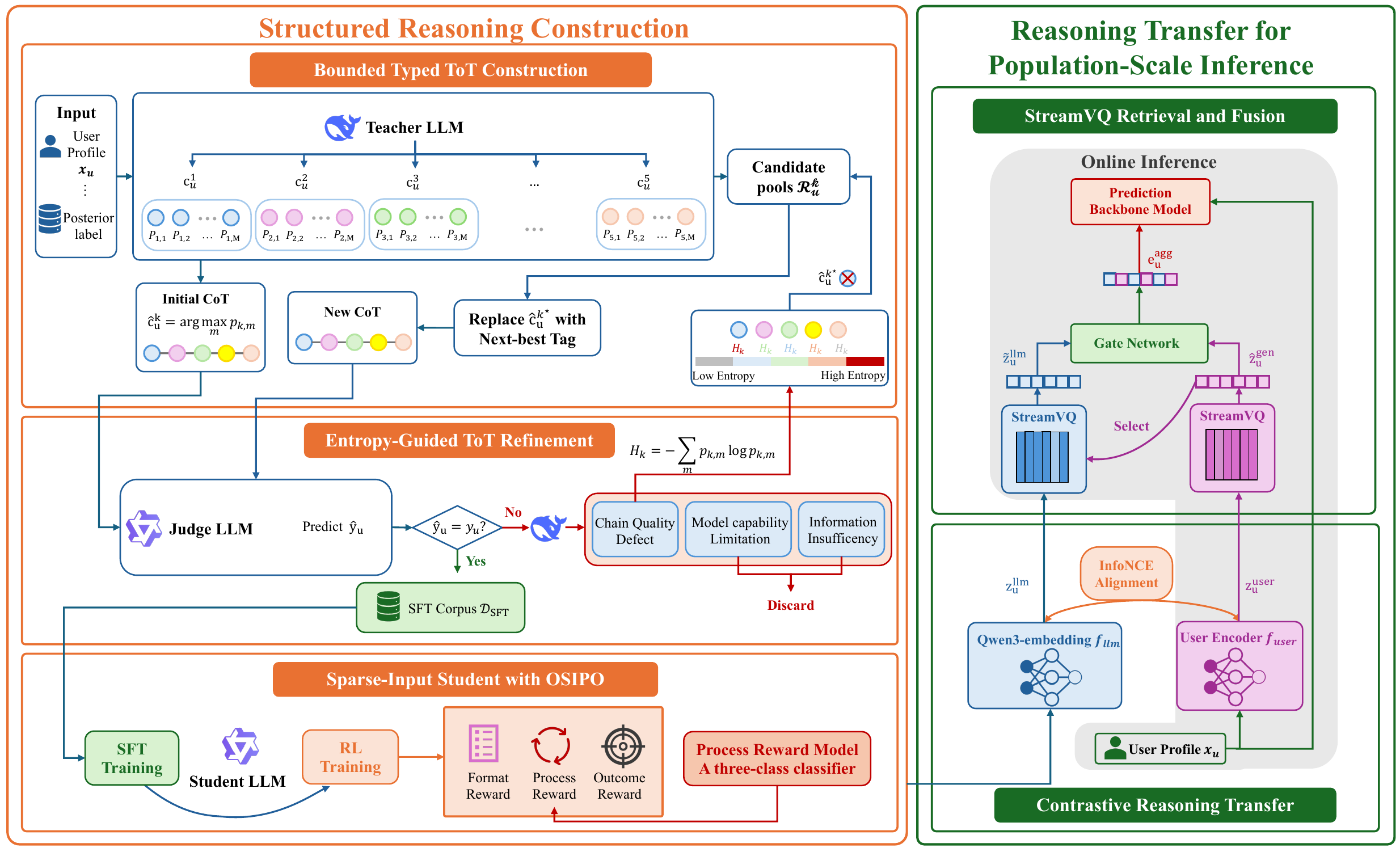}
  \caption{Overview of ScaleToT. The framework has two stages. (1) \emph{Structured Reasoning Construction}: a teacher constructs typed user-state chains with bounded ToT, and a student learns to generate these chains from sparse profiles through SFT and OSIPO. (2) \emph{Reasoning Transfer for Population-Scale Inference}: a lightweight profile encoder maps each sparse profile to a user-specific representation and uses it to retrieve shared reasoning representations from VQ codebooks; a profile-conditioned gate network then combines the user-specific and retrieved representations for prediction. In the advertising LTV instantiation, each chain contains $K{=}5$ lifecycle fields.}
  \label{fig:framework}
\end{figure*}

ScaleToT addresses unreliable reasoning from sparse profiles and the cost of scaling LLM inference to billions of users. For reliability, a teacher uses optional training-only privileged context and bounded ToT to construct typed user-state chains, preserving alternatives and locally revising uncertain fields; a sparse-input student learns these chains with segment-aware feedback. For scalability, the student's reasoning representations are quantized into shared VQ prototypes retrievable from user-specific profile representations. This allows the broader population to fuse profile and retrieved reasoning representations without chain generation or LLM inference.

\subsection{Structured Reasoning Construction}

\subsubsection{Bounded Typed ToT Construction}

ScaleToT represents prediction as inference over $K$ typed fields. Bounded search within $\{\mathcal{T}_k\}_{k=1}^{K}$ replaces one opaque LLM judgment with candidate states that can be compared, revised, and supervised. In the LTV instantiation, the five fields describe user group, churn reason, return motivation, interest preference, and behavior pattern.

Sequential CoT lets early errors distort later fields. Because each field has a finite taxonomy, we instead evaluate all admissible tags independently and retain alternatives for local revision. For field $k$:
\begin{equation}\label{eq:cand}
  \mathcal{C}_u^{k} \;=\; \bigl\{(t_{k,m},\,p_{u,k,m})\bigr\}_{m=1}^{|\mathcal{T}_k|}\,, \quad t_{k,m}\in\mathcal{T}_k\,,
\end{equation}
where $p_{u,k,m}$ is the constrained-decoding probability of tag $t_{k,m}$ and sums to one over $\mathcal{T}_k$. The highest-probability index $m_u^{k\star}=\arg\max_m p_{u,k,m}$ gives the initial value $\hat c_u^k=t_{k,m_u^{k\star}}$; $\mathcal{R}_u^k$ stores the other tags in decreasing probability order. After independent selection prevents cross-field propagation, the teacher serializes the fields into a fixed-order reasoning chain for student supervision.

Sparse inputs may not determine every field. The teacher may therefore use privileged context $p_u$ to construct outcome-consistent supervision for training instances. This context never enters the student, input encoder, codebook query, held-out data, or inference request; all transferred representations derive from student outputs. In advertising, $p_u$ contains conversion context and post-return feedback and statistics.

\subsubsection{Entropy-Guided ToT Refinement}
Parallel generation prevents cross-field propagation but does not ensure that the assembled chain supports the outcome. A frozen judge LLM first tests each chain, and a teacher diagnoses rejected cases. We refine only chain-quality failures; cases attributed to insufficient information or judge limitations are discarded rather than forced into an explanation.

When a chain is rejected, we must decide which field to revise, but no field-level label identifies the error. We therefore measure the teacher's uncertainty from the candidate probabilities: similar probabilities across multiple tags indicate low confidence in the selected tag. Because fields may contain different numbers of tags, we use normalized entropy to compare their uncertainty:
\begin{equation}\label{eq:entropy}
  \begin{aligned}
    \bar{H}_u^k&=-\frac{1}{\log|\mathcal{T}_k|}\sum_{t\in\mathcal{T}_k}p_{u,k}(t)\log p_{u,k}(t),\\
    k^\star&=\arg\max_{k\in\{1,\ldots,K\}}\bar{H}_u^k.
  \end{aligned}
\end{equation}
Here $\bar{H}_u^k\in[0,1]$ is the normalized entropy, $p_{u,k}(t)$ the probability of tag $t$, and $k^\star$ the highest-entropy field. Entropy ranks revision priority rather than estimating calibrated error. We replace $\hat c_u^{k^\star}$ with the next candidate in $\mathcal{R}_u^{k^\star}$, reconstruct the chain, and repeat until judge acceptance or pool exhaustion. One-field-at-a-time revision preserves the other states, and a chain is accepted when it supports the observed outcome.

\subsubsection{Sparse-Input Student with OSIPO}
\begin{figure}[h]
  \centering
  \includegraphics[width=\linewidth]{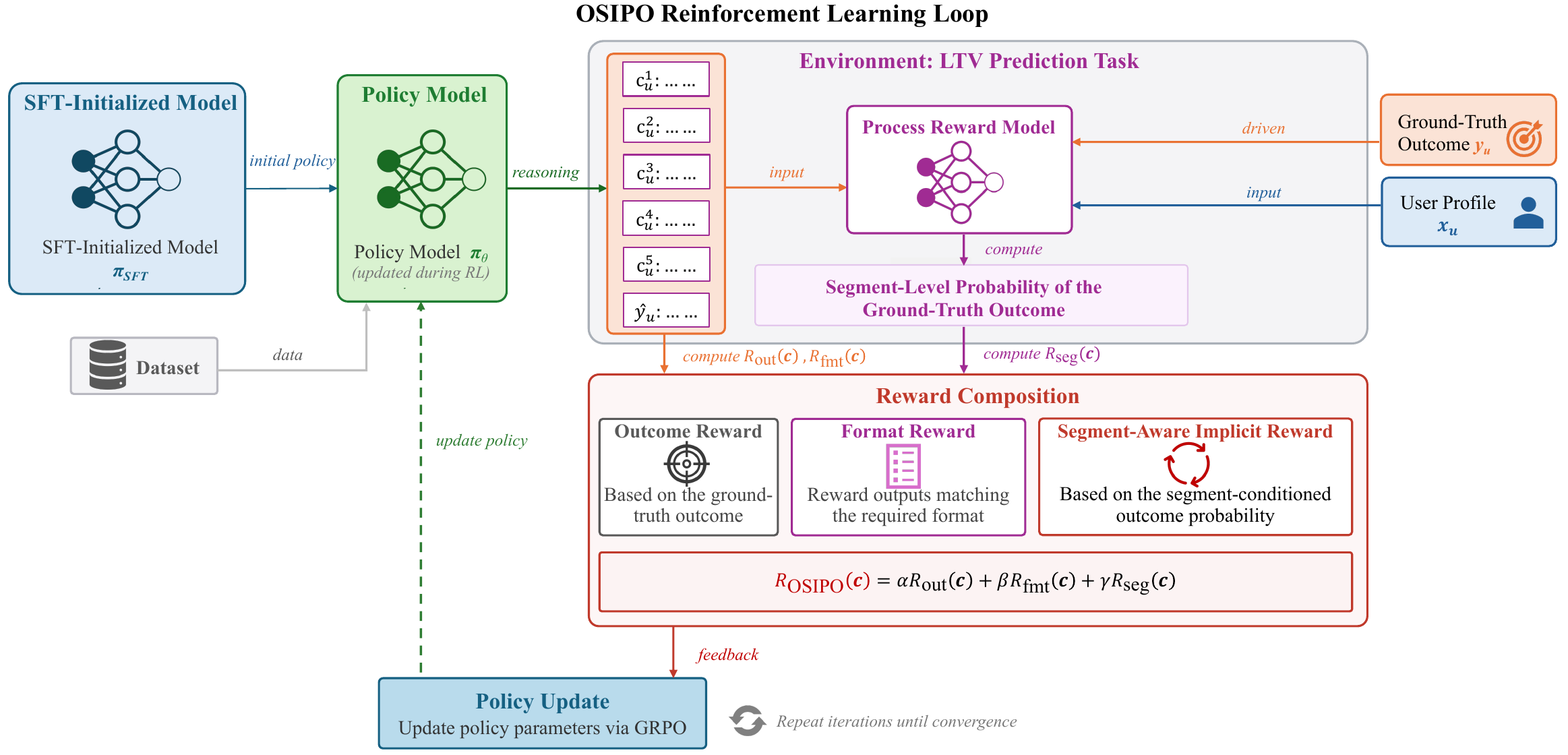}
  \caption{OSIPO reinforcement learning process. An SFT-initialized policy generates structured reasoning chains from sparse user profiles. A process reward model scores how strongly each typed segment supports the ground-truth outcome, and the resulting segment-level reward is combined with outcome and format rewards. GRPO then computes group-relative advantages from the combined reward and updates the policy.}
  \label{fig:rl}
\end{figure}
The next stage transfers embeddings of student-generated chains to the full population. Consequently, the student must preserve meaningful intermediate fields rather than merely predict the correct final label. An outcome reward cannot enforce this requirement: multiple chains can lead to the same prediction, while a single scalar reward neither identifies an inconsistent field nor indicates which part of an incorrect chain should be revised. Format checking only guarantees the presence of the required fields. We therefore introduce OSIPO. In this name, outcome-driven indicates that supervision ultimately comes from $y_u$, segment-aware indicates that credit is conditioned on each typed chain field and its position, and implicit reward indicates that this credit is inferred from outcome compatibility rather than observed segment labels.

The process reward is obtained by training a lightweight PRM on SFT chains to score each generated field's compatibility with the observed outcome, then averaging the field scores:
\begin{equation}\label{eq:rseg}
  R_{\mathrm{seg}}(\bm{c}) = \frac{1}{K}\sum_{k=1}^{K}\Pr_{\mathrm{PRM}}\!\bigl(y_u\mid \bm{x}_u, c_u^{k}, k\bigr)\,,
\end{equation}
where $\Pr_{\mathrm{PRM}}(y_u\mid\bm{x}_u,c_u^k,k)$ scores the observed target for segment $k$. Averaging yields a chain-level reward while preserving segment-aware credit; the score measures target compatibility rather than segment-level correctness.

We combine the implicit segment reward with chain-level objectives:
\begin{equation}\label{eq:reward}
  R_{\mathrm{OSIPO}}(\bm{c}) \;=\; \alpha\,R_{\mathrm{out}}(\bm{c}) \;+\; \beta\,R_{\mathrm{fmt}}(\bm{c}) \;+\; \gamma\,R_{\mathrm{seg}}(\bm{c})\,,
\end{equation}
Here $R_{\mathrm{out}}$ measures target agreement, $R_{\mathrm{fmt}}$ structural compliance, and $\alpha,\beta,\gamma$ weight the three signals. SFT teaches field structure and tag semantics; OSIPO then uses GRPO~\citep{shao2024deepseekmath} to optimize $R_{\mathrm{OSIPO}}$. Thus, OSIPO changes the credit signal for structured chains rather than the underlying group-relative optimizer. The trained student generates $\tilde{\bm c}_u=\pi_\theta(\bm x_u)$ from sparse input alone.

\subsection{Reasoning Transfer for Population-Scale Inference}

Although the student can reproduce teacher reasoning from sparse profiles, running it for billions of users remains too costly.  We therefore ask whether its reasoning can be reused without generating a chain for every user. To answer this question, we generate chains only for a sampled subset and use them to learn how reasoning representations can be recovered directly from sparse profiles. On this subset, contrastive alignment trains a lightweight encoder to map each profile toward its student-chain representation. Because this mapping is noisy under sparse inputs, VQ codebooks organize recurring reasoning patterns into stable, shared prototypes that can be retrieved from a profile alone. Because a shared prototype may not fit every user, a profile-conditioned gate combines the retrieved reasoning representation with the user's own profile representation. The remaining users thus require neither chain generation nor chain encoding.

\subsubsection{Contrastive Reasoning Transfer}

To retrieve reasoning without generating a chain, a user's profile representation must identify the reasoning pattern associated with that user. However, profile and chain representations come from different inputs and are not directly comparable. For each sampled user $u\in\mathcal{U}_{\mathrm{LLM}}$, the input encoder produces $\bm z_u^{\mathrm{user}}=f_{\mathrm{user}}(\bm x_u)$, while the fixed embedding function produces the chain representation $\bm z_u^{\mathrm{llm}}=f_{\mathrm{llm}}(\tilde{\bm c}_u)$. We use the information noise-contrastive estimation (InfoNCE) loss to bring representations of the same user closer and separate those of different users:
\begin{equation}\label{eq:align}
  \mathcal{L}_{\mathrm{align}} = -\frac{1}{B}\sum_{u=1}^{B}\log\frac{\exp\!\bigl(\mathrm{sim}(\bm{z}_u^{\mathrm{user}},\bm{z}_u^{\mathrm{llm}})/\tau\bigr)}{\sum_{v=1}^{B}\exp\!\bigl(\mathrm{sim}(\bm{z}_u^{\mathrm{user}},\bm{z}_v^{\mathrm{llm}})/\tau\bigr)}\,,
\end{equation}
where $\mathrm{sim}$ is cosine similarity, $\tau$ the temperature, and $B$ the batch size. For profile $u$, its own chain is the positive and the other $B-1$ chains are negatives. After training, $f_{\mathrm{user}}$ maps a sparse profile into the aligned space, allowing it to query reasoning prototypes without first generating a chain.

\subsubsection{StreamVQ Retrieval and Fusion}

Contrastive alignment places profile and chain representations in the same space, but a representation derived from a sparse profile may still be noisy. VQ~\citep{van2017neural} improves retrieval stability by mapping similar representations to a finite set of shared prototypes. We maintain two codebooks for the two sources: $\mathbf B^{\mathrm{gen}}$ contains prototypes learned from profile representations, while $\mathbf B^{\mathrm{llm}}$ contains prototypes learned from student-chain representations.

During training, input and chain embeddings update $\mathbf B^{\mathrm{gen}}$ and $\mathbf B^{\mathrm{llm}}$, respectively; StreamVQ~\citep{bin2025real} maintains both through soft assignment and exponential moving averages. The contrastive loss $\mathcal L_{\mathrm{align}}$ and prediction loss $\mathcal L_{\mathrm{cal}}$ jointly train $f_{\mathrm{user}}$. The prediction loss also trains VQ embedding enhancement (VEE), the gate, and $f_\theta$, while $f_{\mathrm{llm}}$ remains fixed.

At inference, $\bm z_u^{\mathrm{user}}=f_{\mathrm{user}}(\bm x_u)$ is quantized by $\mathbf B^{\mathrm{gen}}$ into the denoised query $\hat{\bm z}_u^{\mathrm{gen}}$. This query is stable but carries only input-derived content, so it must read the reasoning side to recover transferred evidence. Because a sparse query rarely matches one prototype exactly, VEE retrieves through soft cross-attention over $\mathbf B^{\mathrm{llm}}$ rather than a hard nearest neighbor:
\begin{equation}\label{eq:vee}
  \tilde{\bm{z}}_u^{\mathrm{llm}}=\mathrm{CrossAttn}\!\bigl(\hat{\bm{z}}_u^{\mathrm{gen}},\;\mathbf{B}^{\mathrm{llm}},\;\mathbf{B}^{\mathrm{llm}}\bigr),
\end{equation}
where $\mathrm{CrossAttn}(\bm q,\mathbf K,\mathbf V)$ uses query $\bm q$, keys $\mathbf K$, and values $\mathbf V$. Here $\hat{\bm z}_u^{\mathrm{gen}}$ queries $\mathbf B^{\mathrm{llm}}$, producing the retrieved representation $\tilde{\bm z}_u^{\mathrm{llm}}$ after projection.

Retrieval can still return reasoning prototypes that do not fit a given user, so the final step commits to neither signal alone and instead weights both per user. Let $\mathcal I=\{\mathrm{gen},\mathrm{llm}\}$ index $\bm s_{u,i}\in\{\hat{\bm z}_u^{\mathrm{gen}},\tilde{\bm z}_u^{\mathrm{llm}}\}$. A profile-conditioned gate assigns:
\begin{equation}\label{eq:gate}
  \mathbf{g}_u = 2\,\sigma\!\bigl(\mathbf{W}_2\,\mathrm{ReLU}(\mathbf{W}_1 \bm{h}_u)\bigr)\in\mathbb{R}^{|\mathcal{I}|}\,,
\end{equation}
\begin{equation}\label{eq:agg}
  \bm{e}_u^{\mathrm{agg}} = \sum_{i\in\mathcal{I}} g_u^{(i)}\bm{s}_{u,i}
\end{equation}
where $\bm h_u$ is the pre-projection encoder state; $\mathbf W_1$ and $\mathbf W_2$ are trainable; $\mathrm{ReLU}$ and $\sigma$ are rectified-linear and sigmoid activations; and $|\mathcal I|=2$. The gate $\mathbf g_u$ weights each path in $\bm e_u^{\mathrm{agg}}$. Its unnormalized range $(0,2)$ allows independent suppression or amplification.

\paragraph{Inference Complexity.}
Let $C_{\mathrm{gen}}$ and $C_{\mathrm{llm}}$ be codebook capacities and $d$ the representation dimension. ScaleToT adds $O((C_{\mathrm{gen}}+C_{\mathrm{llm}})d)$ computation and prototype memory per request, independent of population size, without LLM inference. This excludes the shared production backbone and raw input features.

\section{Experimental Evaluation}\label{sec:eval}
Our experiments answer three questions. First, does ScaleToT improve LTV prediction and reasoning quality over baselines given the same sparse profile inputs? Second, how does each component contribute to performance and scalability? Third, do the offline gains translate into measurable benefits in a billion-scale advertising system? We first describe the experimental setup, then report the overall results, ablation and scalability analyses, and randomized online A/B test.

\subsection{Experimental Setup}
\subsubsection{Dataset and Metrics}
We evaluate ScaleToT on a large industrial dataset from the user-growth system of a major commercial platform. The data cover returning users in an advertising delivery scenario. Users inactive for at least 30 consecutive days are considered low activity, and their activity in the seven days after return defines the LTV label. Training uses 30{,}000 SFT samples and 15{,}000 RL samples, balanced across three ordinal LTV classes. All data are anonymized and aggregated.

We evaluate reasoning quality, LTV prediction, and online impact. Ranking AUC measures how well method outputs scored by a fixed outcome scorer distinguish the ordered LTV classes. The scorer is trained only on the training partition, frozen before evaluation, and shared across methods; it uses no held-out outputs, privileged context, or validation/test labels. Because individual reasoning fields have no ground-truth labels, Ranking AUC evaluates only whether the overall output is predictive of LTV. Embedding Discrimination measures the mean pairwise Euclidean distance between class centroids from fixed Qwen3-Embedding representations; because it is scale dependent and ignores within-class variation, we use it only as a representation diagnostic.

For LTV prediction, we report AUC on a fixed evaluation population with its empirical class distribution. Coverage experiments report absolute AUC, while ablations report percentage-point (pp) gains over a profile-only production backbone. This backbone directly feeds the same user-profile fields into the downstream LTV predictor and does not use any reasoning representation. Online, we report LT30, the cumulative active days within 30 days after RTB delivery, and the predicted-over-clicked ratio (PCOC), where values closer to 1 indicate better calibration. LT30 with its 99\% confidence interval is the confirmatory measure; PCOC and all offline results are descriptive because repeated-run estimates or confidence intervals are unavailable. Exact cohort sizes, dates, and absolute LT30 values are withheld under the proprietary-data agreement.

\subsubsection{Baselines and Implementation Details}
We compare four controlled internal baselines under the same sparse-profile input boundary, LLM backbone, Qwen3-Embedding model, and generation budget:
\begin{itemize}
 \item \textbf{Direct LLM (D-LLM)}: the LLM directly predicts from $\bm{x}_u$ in a single pass, without producing a retained reasoning chain, typed-field decomposition, or closed-set nodes. It serves as the direct profile-only reference.
\item \textbf{Free-Form CoT (FF-CoT)}: the LLM generates free-form chains that are retained without quality screening or closed-set node constraints.
\item \textbf{Quality-Selected CoT (QS-CoT)}: free-form chains pass through the same quality-control pipeline as ScaleToT but use no closed-set taxonomy.
\item \textbf{Sequential CoT (Seq-CoT)}: the same five typed nodes are generated autoregressively, with each node conditioned on its predecessors, under the same consolidation and quality-control pipeline as ScaleToT.
\end{itemize}
These baselines isolate the effects of quality screening, typed decomposition, and parallel node generation within the same sparse-input reasoning pipeline. The fixed production backbone provides the reference for end-to-end LTV gains.

DeepSeek-R1 serves as the teacher, Qwen3-32B as the frozen chain judge, Qwen3-8B as the student reasoner, and Qwen3-0.6B as the PRM backbone. SFT uses a learning rate of $2{\times}10^{-5}$ with cosine decay, a batch size of 32, and 3 epochs. OSIPO uses GRPO with a learning rate of $1{\times}10^{-6}$ and a group size of 8. A frozen Qwen3-Embedding model produces 256-dimensional chain embeddings. The profile encoder $f_{\mathrm{user}}$ is a multilayer perceptron (MLP) with three layers and hidden dimension 512, each VQ codebook contains 1{,}024 entries, and InfoNCE uses $\tau{=}0.07$ with a batch size of 1{,}024.

\subsection{Overall Performance}

\begin{table}[t]
  \centering
  \caption{Reasoning quality comparison on the LTV task. The best and second-best results are in bold and underlined.}\label{tab:cot}
  \begin{tabular}{lcc}
    \toprule
    Method & Rank.\,AUC & Emb.\,Discr. \\
    \midrule
    D-LLM & 0.502 & 0.106 \\
    FF-CoT          & 0.533 & \underline{0.228} \\
    QS-CoT       & 0.562 & 0.147 \\
    Seq-CoT     & \underline{0.589} & 0.115 \\
    ScaleToT           & \textbf{0.615} & \textbf{0.449} \\
    \bottomrule
  \end{tabular}
\end{table}

Table~\ref{tab:cot} evaluates whether generated outputs help distinguish LTV levels. All methods use profiles alone, without evaluation-time labels or privileged information. ScaleToT achieves the highest Ranking AUC of 0.615, ahead of Seq-CoT at 0.589, QS-CoT at 0.562, and FF-CoT at 0.533. This result shows that typed parallel reasoning captures more information related to the LTV target. ScaleToT also reaches the highest Embedding Discrimination of 0.449 among the internal baselines. This metric measures only the distance between class centers, so a higher value does not necessarily imply better LTV prediction. ScaleToT uses a closed set of five fields, so its text is less diverse than free-form chains by design. Because individual fields have no ground-truth labels, these metrics evaluate the usefulness of the complete chain rather than the correctness of each field.
D-LLM achieves a Ranking AUC of 0.502, close to random ranking, showing that a direct LLM output provides little discriminative signal. ScaleToT instead transfers patterns learned across many structured chains and combines them with profile representations. Ranking AUC in Table~\ref{tab:cot} evaluates the generated outputs alone, while downstream AUC evaluates the complete prediction model.

\subsection{Ablation Study}

\begin{table}[t]
  \centering
  \caption{Component ablation at 20\% LLM reasoning coverage on offline dataset. All variants use the same training setup.}\label{tab:ablation}
  \resizebox{\columnwidth}{!}{
  \begin{tabular}{lc}
    \toprule
    Variant &  Downstream AUC \\
    \midrule
    ScaleToT (full)             & \textbf{+1.28}\,pp \\
    \midrule
    w/o Contrastive Reasoning Transfer             & +0.94\,pp \\
    w/o Cross-Attention Retrieval & +0.72\,pp \\
    w/o VQ Codebook Denoising        & +1.05\,pp \\
    w/o Gated Fusion          & +1.10\,pp \\
    \bottomrule
  \end{tabular}}
\end{table}

Table~\ref{tab:ablation} evaluates each module under a fixed training configuration with approximately 20\% reasoning coverage on offline dataset. All ScaleToT variants follow the same sparse-input prediction boundary and use the reasoning-representation path. The profile-only backbone is the Base setting: it feeds the same user-profile fields directly into the downstream LTV predictor, without reasoning representations, retrieval, or fusion, and achieves an AUC of 0.7589. The full model reaches 0.7717 for a gain of 1.28\,pp. Removing Cross-Attention Retrieval causes the largest drop and reduces this gain to 0.72\,pp. Removing Contrastive Reasoning Transfer, VQ Codebook Denoising, and Profile-Conditioned Gated Fusion reduces the gain to 0.94, 1.05, and 1.10\,pp, respectively. These results show the sensitivity of ScaleToT to each module under this configuration.

Entropy-guided refinement reduces average training-chain perplexity under the frozen Qwen3-32B judge from 23.02 to 17.94 (22.1\%). This supports improved evaluator coherence, not field correctness or isolated entropy causality.

\begin{figure}[t]
  \centering
  \includegraphics[width=0.85\columnwidth]{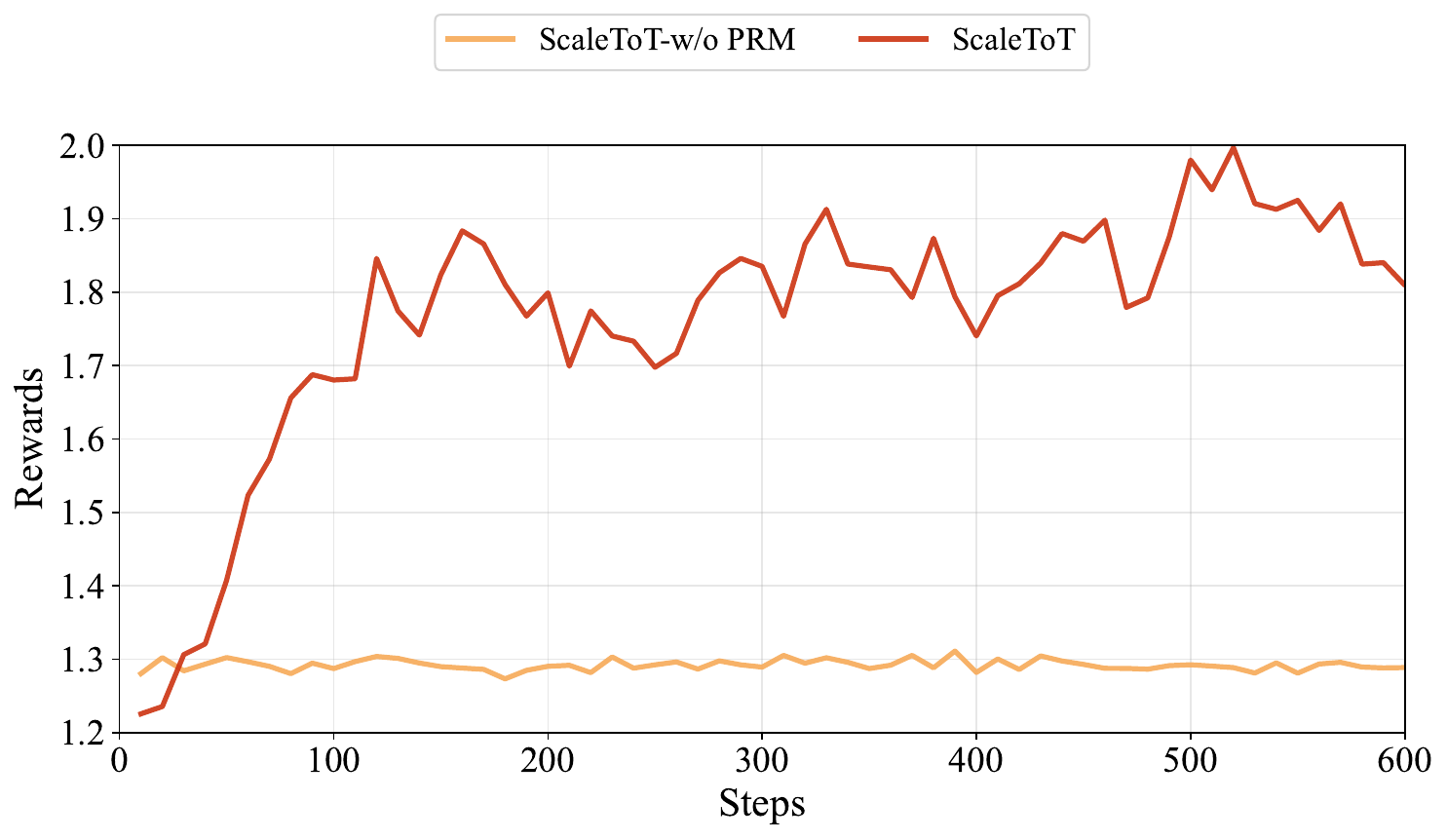}
  \caption{OSIPO training reward over steps with and without the segment-aware implicit reward (optimization diagnostic).}
  \label{fig:reward}
\end{figure}

Fig.~\ref{fig:reward} compares GRPO training with the complete OSIPO reward against training with only outcome and format rewards. With the segment-aware process reward, the training curve rises and stabilizes, while the outcome-and-format reward oscillates and fails to converge. This result indicates that segment-level feedback helps GRPO converge when learning structured chains. Because the two rewards use different scales, their absolute values are not directly comparable.

\subsection{Scalability Analysis}

\begin{figure}[t]
  \centering
  \includegraphics[width=0.9\columnwidth]{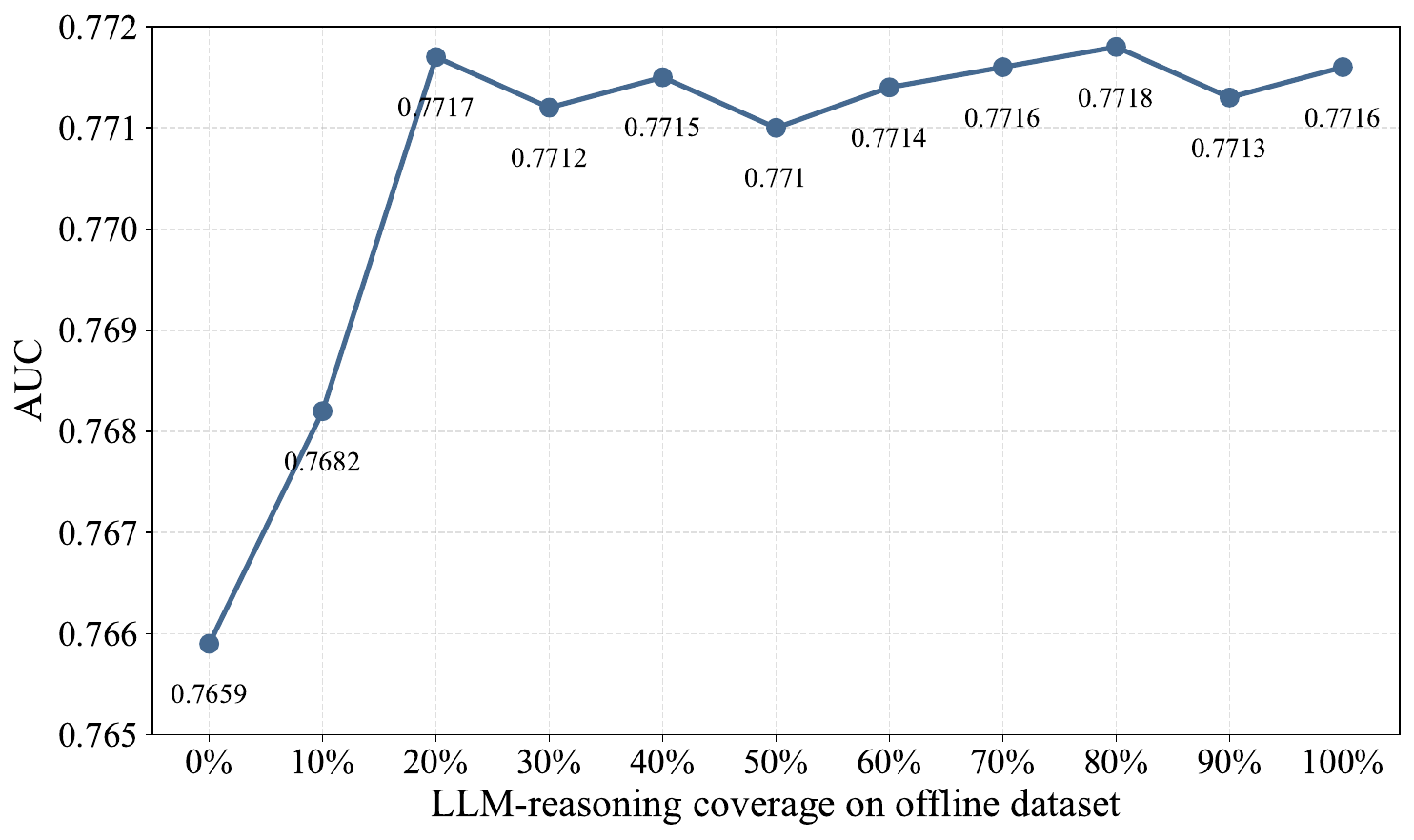}
  \caption{Downstream AUC at different levels of LLM-reasoning coverage on the 366M-user offline dataset. Each point is trained and evaluated at its corresponding offline-dataset coverage level.}
  \label{fig:coverage_auc}
\end{figure}

To examine the scalability of reasoning transfer, Fig.~\ref{fig:coverage_auc} studies how downstream AUC depends on LLM-reasoning coverage, defined as the fraction of users in the 366M-user offline dataset that have an LLM-generated reasoning chain while the rest are inferred by profile-based retrieval and fusion. We measure this relationship on the offline dataset and retrain the model at each coverage level. At 0\% coverage, no offline users receive LLM-generated chains for training, so all users are inferred through VQ-denoised profile representations with gated fusion; ScaleToT reaches 0.7659 AUC. AUC rises to 0.7717 at 20\% coverage and then remains stable, suggesting that additional LLM-reasoning coverage provides little extra benefit after the main diversity of reasoning states is covered. Since 20\% of the 366M-user offline dataset corresponds to 73.2 million LLM-reasoning users, this operating point covers only 7.32\% of the roughly one-billion potential user population. In production, the reasoning model generates chains offline for about 73.2 million users, while all remaining users use profile-based retrieval and fusion without LLM inference.

\subsection{Online A/B Test}

\begin{table}[t]
  \centering
  \caption{Online A/B test with a 7-day allocation window and 30-day post-RTB LT30 follow-up.}\label{tab:ab}
  \begin{tabular}{lr}
    \toprule
    Metric & Result \\
    \midrule
    LT30      & +6.738\% \\
    LT30 confidence      & 99\% \\
    LT30 CI      & [1.54\%, 10.49\%] \\
    PCOC                   & 1.076 $\rightarrow$ 1.034 \\
    \bottomrule
  \end{tabular}
\end{table}

We ran a randomized seven-day A/B test and assigned 5\% of low-activity traffic to ScaleToT. LT30 was measured for both groups over the same 30 days after RTB delivery. As shown in Table~\ref{tab:ab}, ScaleToT increases LT30 by 6.738\% over the production backbone without online LLM inference. The 99\% confidence interval ranges from 1.54\% to 10.49\%. PCOC moves from 1.076 to 1.034, closer to the ideal value of 1. The test compares the complete systems.

\section{Conclusion}\label{sec:conclusion}

We presented ScaleToT, a framework that makes structured LLM reasoning more reliable under sparse profiles and reusable at population scale. ScaleToT constructs typed user-state chains with bounded ToT and entropy-guided local revision, then trains a sparse-input student using OSIPO's outcome-driven, segment-aware rewards. It further transfers student reasoning into profile-retrievable VQ prototypes, enabling prediction without per-user chain generation or LLM inference. In a billion-scale advertising deployment for LTV prediction, offline chains were generated for 7.32\% of the potential population, while a randomized online A/B test increased LT30 by 6.738\%. Our evaluation is limited to this domain, and validating ScaleToT in other sparse-input tasks remains future work.

\section*{AI-Generated Content Acknowledgement}
We utilized generative AI tools (\eg ChatGPT) exclusively for grammatical and stylistic refinement of the manuscript. These tools were not used for content creation, data analysis, or experimental design.

\bibliographystyle{icml2026}
\bibliography{references}

\clearpage
\appendix
\section{Extended Preliminaries}\label{app:prelim}
We review the three technical foundations used by ScaleToT, namely structured LLM reasoning, RL for reasoning, and VQ for scalable representation retrieval.

\subsection{Structured Reasoning}
CoT prompting~\citep{wei2022chain} elicits a free-form reasoning trace $r=(r_1,\dots,r_T)$ generated in a single autoregressive pass. This formulation improves interpretability but commits to earlier steps without explicit revision. ToT reasoning~\citep{yao2023tree} generalizes CoT by modeling reasoning as a search over a state space of partial thoughts. Formally, ToT maintains a set of candidate states $\mathcal{S}=\{s_1,\dots,s_b\}$, where each state $s$ represents a partial reasoning path. At each step, a generator $G(s)$ proposes successor states and a value function $V(s)$ evaluates their promise, enabling the search procedure to expand, prune, or backtrack. ScaleToT borrows the ToT idea of generating multiple candidates and revising uncertain reasoning steps, but it does not perform open-ended tree search. Instead, it instantiates ToT as bounded typed search: it fixes the reasoning structure as a finite set of task-defined typed fields and applies candidate generation and entropy-guided revision only within these fields, making the process easier to control and score.

\subsection{RL for Reasoning}
A reasoning model can be formulated as a policy $\pi_\theta(\bm{c}\mid\bm{x})$ that generates a structured chain $\bm{c}$ conditioned on an input $\bm{x}$. GRPO~\citep{shao2024deepseekmath} optimizes such policies by sampling a group of candidate chains and computing relative advantages within the group, avoiding the need for a separate value model. However, an outcome reward defined only on the final prediction cannot identify which segment in a structured chain supports or conflicts with that outcome. Process reward models~\citep{lightman2024let} address this limitation by assigning credit to intermediate reasoning steps. OSIPO specializes this principle to typed chains: it infers segment-aware credit from the observed outcome while retaining GRPO as the underlying optimizer.

\subsection{VQ}
VQ~\citep{van2017neural} maps a continuous representation $\bm{r}$ to its nearest entry in a codebook $\mathbf{B}=\{\bm{b}_1,\ldots,\bm{b}_C\}$. This discretization converts dense representations into reusable prototypes and enables efficient retrieval; by collapsing nearby continuous vectors onto a shared entry, it also denoises and regularizes representations derived from sparse, noisy inputs. Exponential-moving-average updates provide a lightweight way to maintain codebook prototypes without repeated full reconstruction. In ScaleToT, these properties are used to denoise profile-derived representations and to store and retrieve generalized and LLM reasoning representations for large-scale inference.

\section{Training-Only Privileged Context, Leakage Prevention, and Algorithms}\label{app:method}

\subsection{Training-Only Privileged Context and the Prediction-Time Boundary}
To make the student learn transferable reasoning rather than memorize chains, we introduce an asymmetric teacher--student setting with a strict prediction-time boundary. Training-only privileged context $p_u$, including hindsight observations after user return, is used only by the teacher LLM when constructing structured CoT annotations for training users; it is never an input to the student reasoner, the profile encoder, the codebook query, any held-out validation/test instance, or the inference request. In the advertising instantiation, the privileged context contributing to fields $c_u^{3}$ (conversion materials and contextual information), $c_u^{4}$ (post-conversion engagement feedback), and $c_u^{5}$ (behavioral statistics within seven days after user return) is observed only after the prediction cutoff. Although the observation window of $c_u^{5}$ overlaps with the label window, it is used only as a privileged teacher-side annotation signal and is never exposed to the student model or inference pipeline.

The offline supervision is label-informed: the teacher-side judge uses $y_u$ to retain or refine outcome-consistent chains for training users. This is privileged-information distillation rather than label-independent chain construction. The reasoning codebook is fitted only on the training partition using embeddings of student-generated chains $\tilde{\bm c}_u=\pi_\theta(\bm x_u)$, and no privileged signal, label-window statistic, or teacher annotation from validation/test users is used for codebook updating, prompt selection, or evaluation; all codebooks and parameters are frozen before held-out evaluation. The student receives only the pre-cutoff sparse input $\bm{x}_u$, so the prediction-time boundary rules out input leakage during inference. It does not remove label information from training-corpus construction, establish factual correctness of the field tags, or isolate language reasoning from a supervision-equivalent non-linguistic target. The reported ablations therefore characterize the complete pipeline and its internal variants rather than provide that causal comparison.

\subsection{Structured Reasoning Construction Algorithm}
Algorithm~\ref{alg:chain} formalizes the entropy-ordered revision procedure. In Stage 1 (Lines 1--4), each finite candidate distribution forms a candidate set and pool. In Stage 2 (Lines 5--12), the judge applies the outcome-conditioned admission rule; for a refinable chain, the method computes per-field normalized entropy and selects the highest-entropy field. Normalizing each entropy by $\log|\mathcal{T}_k|$ makes the scores comparable across fields with different candidate-set cardinalities, so uniform taxonomies are not required. This is a deterministic search heuristic, not a calibrated estimate that the selected field is factually wrong. Stage 3 substitutes that field with the next-best tag until judge agreement or pool exhaustion.

\begin{algorithm}[H]
\caption{Entropy-Guided Structured Reasoning Construction}
\label{alg:chain}
\textbf{Input}: Training instance with sparse input $\bm{x}_u$, privileged context $p_u$, and observed outcome label $y_u$; typed candidate sets $\{\mathcal{T}_k\}_{k=1}^{K}$; teacher LLM $\mathcal{G}$, judge LLM $\mathcal{J}$, analyzer LLM $\mathcal{A}$; chain length $K$\\
\textbf{Output}: Structured supervision $\bm{c}_u$, or $\varnothing$ if the instance is discarded
\begin{algorithmic}[1]
\FOR{$k = 1, 2, \dots, K$}
    \STATE Obtain the closed-set predictive distribution over the $|\mathcal{T}_k|$ tags with the teacher $\mathcal{G}(\bm{x}_u, p_u)$ (offline annotation only) and form the per-field candidate set $\mathcal{C}_u^{k}$.
    \STATE Select the initial field value $\hat{c}_u^{k}$ and form the candidate pool $\mathcal{R}_u^{k}$.
\ENDFOR
\STATE Consolidate the per-field selections into the chain narrative $\bm{c}_u$.
\STATE Compute the judge prediction $\hat{y}_u = \mathcal{J}(\bm{x}_u,\bm{c}_u)$.
\IF{$\hat{y}_u = y_u$}
    \STATE \textbf{return} $\bm{c}_u$
\ELSE
    \STATE Perform root-cause analysis with $\mathcal{A}$ to classify the cause of disagreement.
    \IF{cause $\in$ \{information insufficiency, capability limitation\}}
        \STATE \textbf{return} $\varnothing$
    \ELSE
        \STATE Compute the per-field normalized predictive entropy $\bar{H}_k$.
        \STATE Identify the weakest field $c_u^{k^{\star}}$ by $k^{\star}=\argmax_k \bar{H}_k$.
        \WHILE{$\mathcal{R}_u^{k^{\star}} \neq \emptyset$}
            \STATE Replace $\hat{c}_u^{k^{\star}}$ with the next-best entry of $\mathcal{R}_u^{k^{\star}}$ and update $\mathcal{R}_u^{k^{\star}}$.
            \STATE Reconsolidate $\bm{c}_u$ and recompute $\hat{y}_u = \mathcal{J}(\bm{x}_u,\bm{c}_u)$.
            \IF{$\hat{y}_u = y_u$}
                \STATE \textbf{return} $\bm{c}_u$
            \ENDIF
        \ENDWHILE
        \STATE \textbf{return} $\varnothing$
    \ENDIF
\ENDIF
\end{algorithmic}
\end{algorithm}

\subsection{Population-Scale Inference Algorithm}
Algorithm~\ref{alg:inference} summarizes prototype-based reasoning transfer and inference. During training, student-generated chains are embedded by $f_{\mathrm{llm}}$, profiles are encoded by $f_{\mathrm{user}}$, the two views are aligned by contrastive learning, and StreamVQ updates $\mathbf{B}^{\mathrm{gen}}$ and $\mathbf{B}^{\mathrm{llm}}$ for clustering and denoising. After training, all transfer-path parameters and codebooks are frozen. During frozen inference, the profile is encoded and quantized, VEE retrieves and enhances the closest reasoning prototypes from $\mathbf{B}^{\mathrm{llm}}$, the gate selects between the VQ-denoised representation and the retrieved reasoning representation, and the fused feature is concatenated with the raw profile for prediction, without LLM inference or codebook updates.

\begin{algorithm}[H]
\caption{Reasoning Transfer for Population-Scale Inference}
\label{alg:inference}
\textbf{Input}: instances $\mathcal{U}_{\mathrm{sp}}$, LLM-processed subset $\mathcal{U}_{\mathrm{LLM}}$, sparse input $\bm{x}_u$, student $\pi_\theta$, reasoning embedding function $f_{\mathrm{llm}}$, user encoder $f_{\mathrm{user}}$, codebooks $\mathbf{B}^{\mathrm{gen}},\mathbf{B}^{\mathrm{llm}}$, gate parameters $\mathbf{W}_1,\mathbf{W}_2$, backbone $f_\theta$\\
\textbf{Output}: task prediction $\Pr(y_u \mid \bm{x}_u)$
\begin{algorithmic}[1]
\STATE \COMMENT{Training}
\FOR{each minibatch $\mathcal{B}\subseteq\mathcal{U}_{\mathrm{sp}}$}
  \STATE Encode profiles $\bm{z}_u^{\mathrm{user}}=f_{\mathrm{user}}(\bm{x}_u)$.
  \STATE Generate student chains $\tilde{\bm{c}}_u=\pi_\theta(\bm{x}_u)$ and embed them with $f_{\mathrm{llm}}$ to obtain $\bm{z}_u^{\mathrm{llm}}$ on $\mathcal{B}\cap\mathcal{U}_{\mathrm{LLM}}$.
  \STATE Align $\bm{z}_u^{\mathrm{user}}$ with $\bm{z}_u^{\mathrm{llm}}$ by the InfoNCE loss $\mathcal{L}_{\mathrm{align}}$.
  \STATE Quantize $\bm{z}_u^{\mathrm{user}}$ and $\bm{z}_u^{\mathrm{llm}}$ to update $\mathbf{B}^{\mathrm{gen}}$ and $\mathbf{B}^{\mathrm{llm}}$ by StreamVQ exponential moving average.
  \STATE Retrieve $\tilde{\bm{z}}_u^{\mathrm{llm}}$ from $\mathbf{B}^{\mathrm{llm}}$ by VEE and fuse $\hat{\bm{z}}_u^{\mathrm{gen}}$ with $\tilde{\bm{z}}_u^{\mathrm{llm}}$ by the gate into $\bm{e}_u^{\mathrm{agg}}$.
  \STATE Predict $f_\theta(\bm{x}_u,\bm{e}_u^{\mathrm{agg}})$ and update all parameters by the task, alignment, and quantization losses.
\ENDFOR
\STATE Freeze $f_{\mathrm{user}}$, $f_\theta$, $\mathbf{B}^{\mathrm{gen}}$, $\mathbf{B}^{\mathrm{llm}}$, $\mathbf{W}_1$, and $\mathbf{W}_2$.
\STATE \COMMENT{Frozen prediction path: without LLM inference or codebook updates}
\FOR{$u \in \mathcal{U}_{\mathrm{sp}}$}
  \STATE Encode the profile and quantize it to $\hat{\bm{z}}_u^{\mathrm{gen}}$ by $\mathbf{B}^{\mathrm{gen}}$.
  \STATE Retrieve the enhanced reasoning representation $\tilde{\bm{z}}_u^{\mathrm{llm}}$ from $\mathbf{B}^{\mathrm{llm}}$ by VEE cross-attention.
  \STATE Compute the per-user gate $\mathbf{g}_u$ from the profile-side context vector $\bm{h}_u$.
  \STATE Fuse the signals into $\bm{e}_u^{\mathrm{agg}}$ by gated summation.
  \STATE Predict $\Pr(y_u \mid \bm{x}_u)=f_\theta(\bm{x}_u, \bm{e}_u^{\mathrm{agg}})$.
\ENDFOR
\STATE \textbf{return} $\{\Pr(y_u \mid \bm{x}_u) \}$
\end{algorithmic}
\end{algorithm}

\section{Dataset Statistics and Implementation Details}\label{app:dataset}

Table~\ref{tab:dataset} summarizes the dataset statistics of the industrial dataset from the user growth system of a major commercial platform, covering returning users in an advertising delivery scenario. All user data is anonymized and aggregated in accordance with privacy requirements.

\begin{table}[h]
  \centering
  \caption{Dataset statistics.}\label{tab:dataset}
  \setlength{\tabcolsep}{2.5pt}
  \resizebox{\columnwidth}{!}{
  \begin{tabular}{lr}
    \toprule
    Statistic & Value \\
    \midrule
    Low-activity criterion & $\geq 30$ consecutive inactive days \\
    LTV label window & 7 days after return \\
    SFT training samples & 30{,}000 \\
    RL training samples & 15{,}000 \\
    LTV activity-intensity classes & High, Medium, and Low \\
    Training-set class ratio & uniform (33.3\% each) \\
    \bottomrule
  \end{tabular}}
\end{table}

The reasoner backbone is Qwen3-8B. SFT uses learning rate $2{\times}10^{-5}$, cosine decay, batch size 32, and 3 epochs; OSIPO uses GRPO with learning rate $1{\times}10^{-6}$ and group size 8. For representation learning, the LLM side uses frozen Qwen3-Embedding to produce 256-dim chain embeddings, while the user encoder $f_{\mathrm{user}}$ is a 3-layer MLP with hidden dimension 512 and GELU activation; each VQ codebook has 1{,}024 entries, and InfoNCE uses $\tau{=}0.07$ and batch size $1{,}024$. SFT and OSIPO run on eight A100-80G GPUs. Traffic is randomly allocated during a seven-day window, with 5\% of low-activity traffic assigned to the experiment; LT30 is measured for each allocated cohort over the same 30-day window after RTB delivery. Exact cohort sizes, calendar ranges, and absolute LT30 are withheld under the proprietary-data agreement.

For reasoning-quality evaluation, Ranking AUC ranks every method's output with the same outcome score against the ordered LTV target. It is a task-oriented output diagnostic rather than a node-correctness metric. Semantic Richness is computed with an external jieba Chinese tokenizer as a weighted combination of token diversity, content-word ratio, and $n$-gram lexical diversity. Embedding Discrimination is the mean pairwise Euclidean distance between per-class centroids $\bm{\mu}_y$ of fixed Qwen3-Embedding encodings; it omits within-class dispersion and is treated only as a scale-dependent diagnostic. Mean token-level perplexity compares chains before and after revision and measures textual fit under the evaluator, not factual correctness.

Semantic Richness is reported only as a lexical diagnostic: ScaleToT intentionally optimizes constrained semantic sufficiency over a closed task vocabulary, so its lower lexical diversity reflects reduced free-form variation rather than weaker task-grounded reasoning or reduced downstream discriminability. The diagnostic values are 0.360 for D-LLM, 0.403 for FF-CoT, 0.391 for QS-CoT, 0.400 for Seq-CoT, and 0.365 for ScaleToT. ScaleToT's value is therefore not used as evidence for the main effectiveness claim; the primary paper instead reports the standard ROC-based Ranking AUC and the diagnostic produced by the fixed external Qwen3-Embedding encoder.

\section{Additional Results and Case Study}\label{app:results}

\subsection{Perplexity Reduction from Self-Refinement}

\begin{figure}[t]
  \centering
  \includegraphics[width=\linewidth]{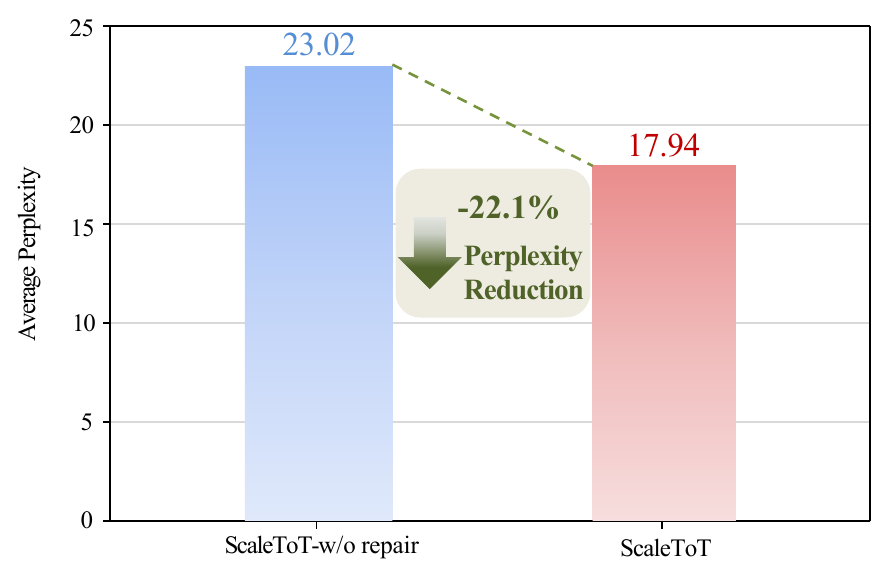}
  \caption{Average token-level perplexity of reasoning chains before and after entropy-guided self-refinement.}
  \label{fig:perplexity}
\end{figure}

Fig.~\ref{fig:perplexity} reports average training-chain perplexity under Qwen3-32B. The value changes from 23.02 before revision to 17.94 afterward. Because revision, candidate order, reconsolidation, and judge admission operate together and no random-node or full-regeneration control is reported, this comparison does not isolate entropy selection. It records that admitted post-revision text has lower evaluator perplexity; it does not establish factual node correctness, calibrated uncertainty, or downstream benefit.

\subsection{Case Study}

\begin{figure*}[b]
  \centering
  \includegraphics[width=\linewidth]{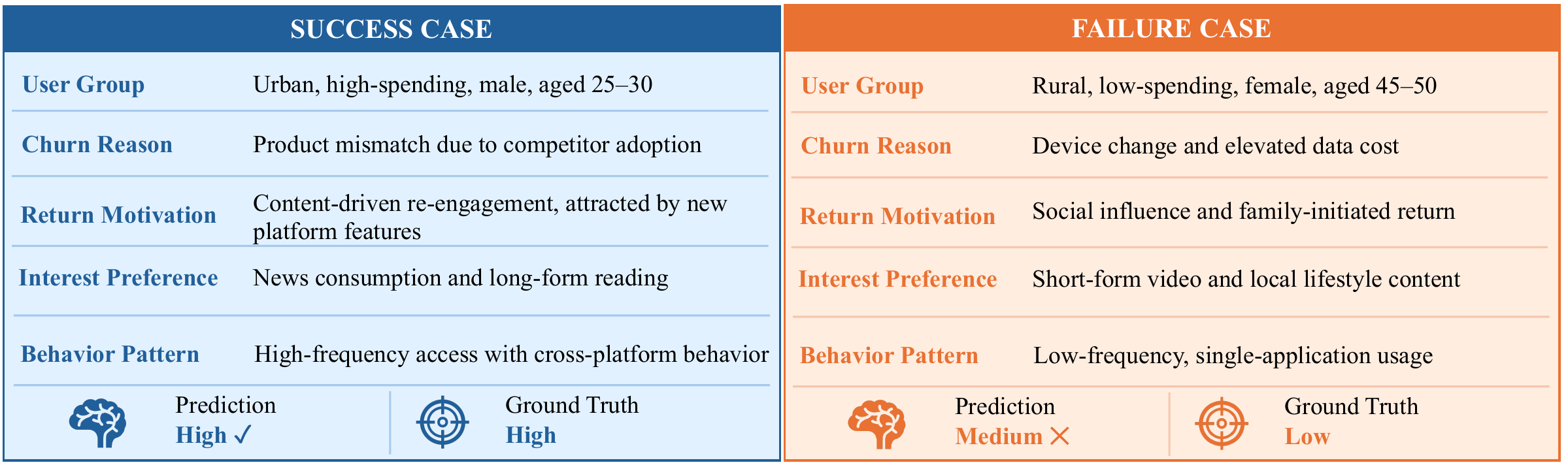}
  \caption{Structured reasoning chains for a correctly predicted user and a mispredicted user.}
  \label{fig:case}
\end{figure*}

Fig.~\ref{fig:case} presents two illustrative returning users, one with a matching final LTV prediction and one without. The examples show the format of the typed nodes and how the implementation orders a local revision by entropy. They are qualitative illustrations rather than representative evidence: the displayed entropies are not calibrated correctness probabilities, and the failure case is compatible with insufficient profile information but cannot establish that explanation on its own.

\end{document}